\documentclass[runningheads]{llncs}
\usepackage[numbers, sort]{natbib}
\usepackage{amsmath, amssymb}
\usepackage{graphicx, algorithm2e}
\usepackage[pagebackref=false,breaklinks=true,letterpaper=true,colorlinks,bookmarks=false]{hyperref}

\usepackage{xcolor}
\newcommand{\todo}[1]{}
\renewcommand{\todo}[1]{{\color{red} \textbf{{#1}}}}

\begin{document}

\title{Multi-modal Graph Fusion for Inductive Disease Classification in Incomplete Datasets}
%
\titlerunning{Multi-modal Graph Fusion for Inductive Classification}  
%
\author{Gerome Vivar\inst{1}\inst{2}\and Hendrik Burwinkel\inst{1} \and Anees Kazi\inst{1} \and Andreas Zwergal\inst{2} \and Nassir Navab\inst{1} \inst{3} \and Seyed-Ahmad Ahmadi\inst{2}} 

%
\authorrunning{Vivar et al.} 
%
%
\institute{Technical University of Munich (TUM), Munich, GER\\
\and 
German Center for Vertigo and Balance Disorders (DSGZ), Ludwig-Maximilians-Universit{\"a}t (LMU), Munich, GER
\and
Whiting School of Engineering, Johns Hopkins University, Baltimore, USA}



\maketitle              
\begin{abstract}
Clinical diagnostic decision making and population-based studies often rely on multi-modal data which is noisy and incomplete. Recently, several works proposed geometric deep learning approaches to solve disease classification, by modeling patients as nodes in a graph, along with graph signal processing of multi-modal features. Many of these approaches are limited by assuming modality- and feature-completeness, and by transductive inference, which requires re-training of the entire model for each new test sample. In this work, we propose a novel inductive graph-based approach that can generalize to out-of-sample patients, despite missing features from entire modalities per patient. We propose multi-modal graph fusion which is trained end-to-end towards node-level classification. We demonstrate the fundamental working principle of this method on a simplified MNIST toy dataset. In experiments on medical data, our method outperforms single static graph approach in multi-modal disease classification.


\end{abstract}
%
\section{Introduction}
\label{sec:intro}
%

Multi-modal, high-dimensional medical data are being collected and stored at an increasing and unprecedented rate in clinical routine. Although computer-aided medical diagnostic systems (CAMD) are already in place at some clinical centers, there is a growing need for CAMD systems which can cope with the rapid increase of medical data that is being collected. A new CAMD systems which can process and analyze big medical data and discover new insights could have the potential to improve diagnosis and deliver better treatments. 
An interesting and promising direction in CAMD has been introduced in recent years in the form of graph-based methods. Here, patients are modeled as nodes in an undirected graph. Adjacencies between patient nodes are modeled by simple similarity measures comparing demographic features and meta-information like clinical scores. Works like \citet{GCN:Parisot17} have shown that feature processing in CAMD approaches can benefit from arranging patients as graph lattices.

\indent \textbf{Related Work}
One of the key ideas of recent methods on deep learning on graphs is the extension of Convolutional Neural Networks (CNNs) to graphs \cite{GCN:Bronstein17}. Here, spectral and non-spectral approaches to graphs have been successfully applied to node-level and graph-level classifications. Under spectral approaches, \citet{Bruna2013} defined the convolution operation in the Fourier domain by computing the eigendecomposition of the Graph Laplacian. Solutions to circumvent the computation complexity of eigendecompositions and localize filters were proposed in \citet{GCN:Defferrard16}. In \citet{Kipf2016}, graph convolutional neural networks were demonstrated to solve semi-supervised classification (GCN), which was applied for CAMD for the first time in \citet{GCN:Parisot17}. Although these spectral approaches have shown state-of-the-art results, they are still limited to the eigenbasis of the graph Laplacian, which changes when new data is introduced. Hence, extensions to inductive learning for classification of out-of-sample data are necessary.
This was addressed recently with non-spectral graph approaches, where convolutions are applied directly on the graph. Here, one of the key ideas is to define an operator which works with arbitrary sized neighborhoods and at the same time has the weight sharing property of CNNs \cite{Velickovic2017}. One such approach is GraphSAGE \cite{Hamilton2017} which operates on the node's local neighborhood. It samples a fixed-size neighborhood of a node and learns to aggregate node information using different aggregation functions, e.g. a LSTM in their best-performing experiment. One issue with this approach is that during inference the network does not have full access to the whole neighborhood. Another non-spectral approach is Graph Attention Networks (GAT) \cite{Velickovic2017} which also works on arbitrary sized neighborhoods, but learns self-attention scores on the full 1-hop neighborhood. Unlike GraphSAGE, all the neighbors of the node are allowed to contribute to the target node in GAT. Their contributions are weighted via a single-layer feed forward neural network. One advantage of these non-spectral approaches is that it can be used for out-of-sample datasets. We can train a model using only the training set and evaluate this model on the held-out test set unlike previous spectral approaches to graphs. We base our work on GAT, due to its consideration of the whole neighborhood, which leads to superior performance over other inductive approaches like GraphSAGE \cite{Velickovic2017}.
\indent \textbf{Contribution.}
In this work, we propose a novel graph neural network architecture designed for CAMD in realistic medical datasets. Our architecture supports (1) out-of-sample, node-based classification in an inductive setting, (2) multi-modal input data, where each modality is handled by a separate processing branch and (3) block-wise incomplete data, modeling missing features from entire modalities which were not acquired due to cost or lack of medical indication. 
Finally, we introduce a graph neural network fusion to aggregate information from multiple modalities in irregular domains such as population graphs.

\section{Materials and Methods}
\label{sec:methods}
\subsection{Dataset and Preprocessing}
To demonstrate our method, we utilize two datasets, TADPOLE\footnote{\url{http://adni.loni.usc.edu}} \cite{Marinescu2018} and a modified version of MNIST. 
On the TADPOLE dataset, the goal is to predict whether a subject with mild cognitive impairment (MCI) will convert to Alzheimer's disease (AD) given their diagnostic parameters at baseline. We select all unique subjects with baseline measurements from ADNI1, ADNIGO, and ADNI2 in the TADPOLE dataset which were diagnosed as MCI including those diagnosed as EMCI and LMCI. We retrospectively use labels of those subjects whose condition progressed to AD within 48 months as pMCI and those whose condition remained stable as sMCI, similarly to Thung et al. \cite{SWMC:Thung16}. The remaining MCI subjects who progressed to AD after month 48 are excluded from this study. We use multi-modal features from MRI, PET, and CSF at baseline, i.e. excluding longitudinal features to fit the block-wise missing setting. We mask modalities to at least have one modality complete and include those subjects who satisfy this criteria. As input to the network we use all numerical features from this dataset and use 3 modalities (MRI, PET, CSF) as 3 different inputs to the multi-modal network.


For a complementary analytic experiment, we adapt the MNIST dataset to fit our problem setting. We perform digit classification, using 10,000 images from the training set, i.e. 1000 instances of every class label. We modify MNIST to mimic a CAMD setting with multi-modal and block-wise (i.e. modality-wise) missing data. To this end, we crop the 28 x 28 images into three non-overlapping images, which simulates a 3-modality setting similar to TADPOLE. We use the lower half of each image as the first modality. We crop the upper-half part of the image into two further parts of unequal size $14 \times 9$ and $14 \times 19$. With this cropping strategy, we simulate three modalities with different discriminative power, covering different locations and percentage of the overall area of the image. As input to the model, we flatten these features to achieve $N \times F$ dimensions, where N is the number of nodes and F is the number of features. To have an incomplete dataset, we randomly drop a predefined percentage (10\% to 50\%) of "modalities" block-wise from the dataset, making sure that there is only one modality missing per node.


%
\subsection{Graph construction}
\label{sec:GraphCons}
Recent graph construction strategies for population based studies have been based on either manual initialization using patients' meta-information (i.e. age and gender) or calculation of certain distance metric on the node features \cite{Parisot2017}. For block-wise missing data we propose to use $M$ independent graphs. We initialize an undirected graph $A$ for each of the $M$ different modalities as $\{A_1, A_2, ..., A_M\}$. As baseline comparisons, we use three strategies to initialize the graph whenever applicable. The first is to use its node feature matrix to calculate its nearest-neighbors (NN-Graph), the second is to initialize using meta-information (MI-Graph) (e.g. age), and the third is to use a fully-connected graph (FC-Graph).


On the TADPOLE dataset, we initialize a graph by connecting subjects to their 10-nearest-neighbors using node features for the NN-Graph. For the MI-Graph, we use risk-factors such as APOE-4, age, and gender to connect patients with the same number of gene risk factors, and increase the weight connection by one, if they have the same gender or age not greater than two years old, as previously done in \cite{Parisot2017}. On MNIST, we create three independent graphs, one for each of the three fake modalities that we generate with non-overlapping crops. Then, using the flattened features of these modalities, we initialize the NN-Graph by connecting each node to its 10-nearest neighbors.


%
\subsection{Graph Attention Network Fusion for Multi-modal and Incomplete Datasets}
In this work, we propose to solve disease classification in incomplete multi-modal dataset with block-wise missing features using GAT with a late fusion approach as illustrated in \ref{fig:Model}. 
A GAT is based on graph attention layers, where each layer takes in a set of node features, $\vec{h} = \{\vec{h}_1, \vec{h}_2, ..., \vec{h}_N\}, \vec{h_i} \in \mathbf{R}^{F}$. Here, $N$ is the number of nodes, and $F$ is the number of features in each node. It outputs a new set of node features, $\vec{h}^{\prime} = \{\vec{h}^{\prime}_{1}, \vec{h}^{\prime}_{2},..., \vec{h}^{\prime}_{N}\}, \vec{h}^{\prime} \in \mathbf{R}^{F^{\prime}}$, $F^{\prime}$ is the output dimension. Given a node feature vector $\vec{h}_i$, a GAT layer forward propagation computes a weighted combination of the neighbors of this node using equation \eqref{eq:GATFProp},

\begin{equation}
\label{eq:GATFProp}
\vec{h}^{\prime}_{i} =  \sigma \bigg(\sum_{j \in {N_i}} \alpha_{i,j} \mathbf{W} \vec{h}_j \bigg)
\end{equation}

where $W \in \mathbf{R}^{F^{\prime} \times F}$ is a weight matrix to linearly transform its neighbor node feature $\vec{h}_j$, scaled by $\alpha_{i,j}$ which are the self-attention scores and $\sigma$ is a non-linear function.

\begin{equation}
\label{eq:GATAttentionScores}
    \alpha_{ij} = \frac{\exp(\text{LeakyReLU}(\vec{a}^T[\mathbf{W}\vec{h}_i \parallel \mathbf{W}\vec{h}_j]))} {\sum_{k \in{N_i}} \exp(\text{LeakyReLU}(\vec{a}^T[\mathbf{W}\vec{h}_i \parallel \mathbf{W}\vec{h}_k]))}
\end{equation}

The attention scores $\alpha_{i,j}$ indicate how important node $j$'s feature is to node $i$.
It is calculated using equation \eqref{eq:GATAttentionScores}, where $\vec{h}_i$ and $\vec{h}_j$ are node features of node $i$ and $j$, $\parallel$ is concatenation, $\vec{a} \in \mathbb{R}^{2F^{\prime}}$ is the attention mechanism's learnable weight vector, exp is the exponential function, LeakyReLU denotes Leaky Rectified Linear Unit, and $k$ is the indices of all the other neighboring nodes of $i$. \citet{Hamilton2017} found that using multiple instances of these self-attention mechanisms stabilizes the learning process. With $K$ instances of self-attention, the output at the final layer of a single GAT layer is then calculated as: 

\begin{figure}[t]
  \centering
  \includegraphics[width=1.0\textwidth]{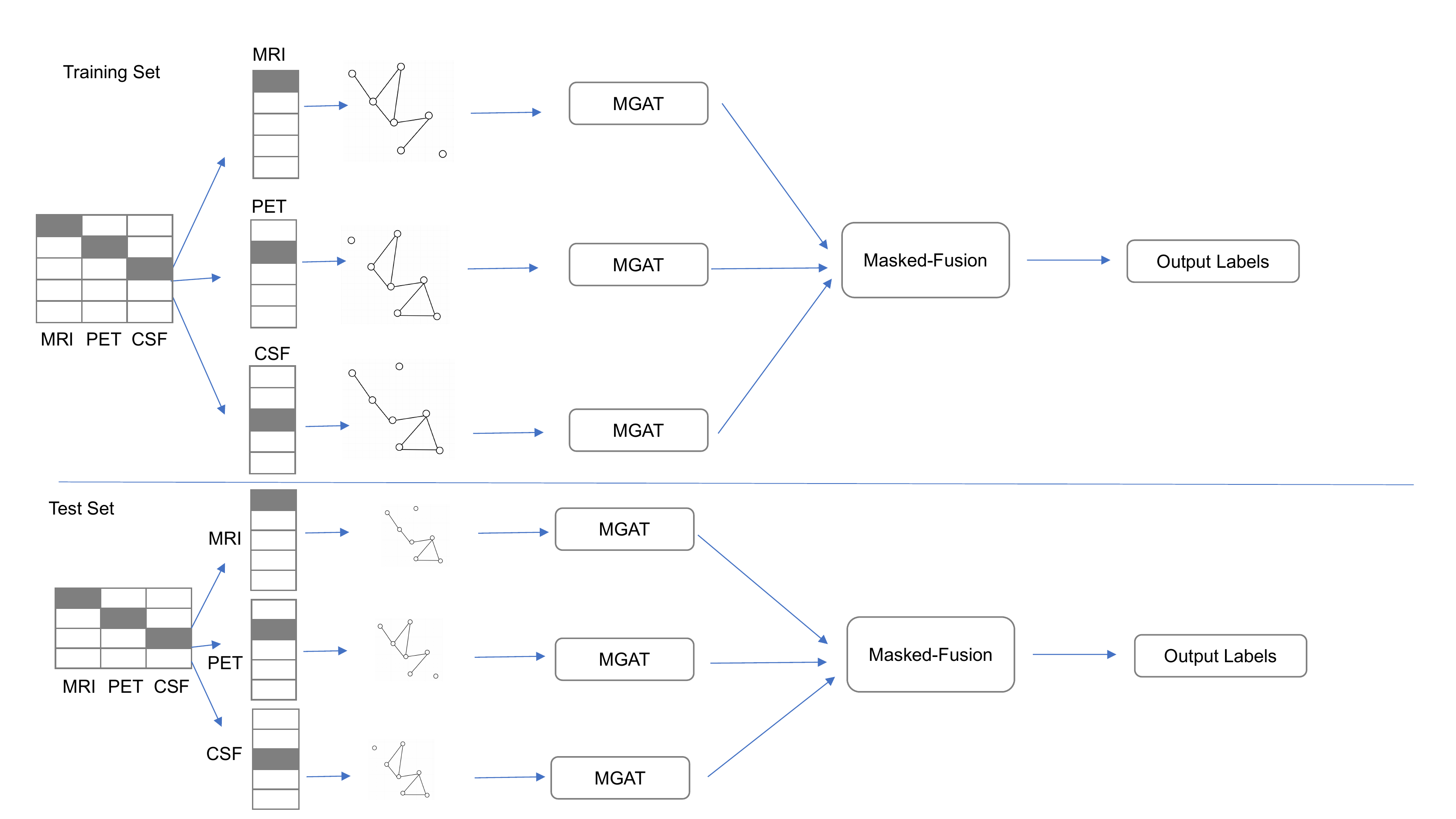}\\
  \caption{Illustration of the overall approach: Given a feature matrix $X$ coming from different modalities and blockwise missing features, 3 independent graph is constructed out of these modalities using its own modality features. A Modified-GAT is used to propagate node-information and a masked fusion is used to output class labels.}
  \label{fig:Model}
\end{figure}


\begin{equation}
\label{eq:GATFPropMean}
\vec{h}^{\prime}_{i} = \sigma \bigg(\frac{1}{K} \sum^{K}_{k = 1} \sum_{j \in {N_i}} \alpha_{i,j}^{k} \mathbf{W}^k \vec{h}_j \bigg)
\end{equation}

Alternatively, to combine these $K$ instances of self-attention, concatenation is also possible by concatenating the outputs of equation \eqref{eq:GATFProp}. In our architecture, we dedicate one instance of a GAT layer to each modality. Given a feature matrix $\mathbf{X}$ of size $N \times F$, we split our dataset into training $X_{train}$ and test $X_{test}$ sets. Using the training set, we train a model that takes in multiple graphs which are coming from the different modalities and fuse unnormalized-log probabilities at the last layer using a masked loss function. Here, we take the masked logits and calculate the mean as input for the sigmoid or softmax function. The network then outputs class labels of size $N_{train} \times C$, where $C$ is the number of classes in a multi-class setting (1-$C$ for binary setting).

In TADPOLE, as illustrated in Figure \ref{fig:Model}, we construct three graphs which are coming from three modalities (MRI, PET, CSF). We initialize the graphs as discussed in \ref{sec:GraphCons}. Given this graph and feature matrix, our GAT-Fusion model takes in all three graphs and learns a model that is able to predict whether a node from a patient will turn into AD or will stay with the MCI condition. To simplify the approach, a node in the graph which is a single patient observation from one modality is disconnected from the graph when there is no measurement done at this modality for this patient. With this approach the network is able to learn end-to-end how to perform prediction even if certain modalities from a patient are not available. At the same time, all information from other modalities coming from other patients can be leveraged and classification can be performed at inference time even for patients which were not included during training, an advantage which transductive methods do not directly have.

 The modified MNIST dataset serves as a complementary experiment to confirm that the method performs on other datasets as well. Since the dataset is complete, it serves also as a testbed to explore certain parameters like the amount of missingness of data.

We use two different GAT stacking approaches which we simply call MGAT as in \ref{fig:Model}. A single MGAT can either be a 2 layer GAT layers (GAT2) or a GAT layer with a multi-layer perceptron (MLP) (GAT1). We only use a single type of layer per graph for every experiment to simplify the approach. In the end, to learn a inductive model we optimize a classification loss term.


%
\subsection{Experimental Setup}
\indent \textbf{Baselines.} As baselines, we use a linear Support Vector Machine (SVC) and Random Forest for standard ML comparison where the missing features are completed with mean imputation. We also use a two layer GAT with mean imputation (GAT-Imp) as another graph neural network baseline. On all baseline experiments, we use a single feature matrix matrix where all modalities are stacked together. In GAT2, we use eight number of heads at the first GAT layer and one head at the 2nd GAT layer. For GAT1, we use the same eight number of heads at the first GAT layer and a MLP at the last layer. To further evaluate every method's performance, we perform stratified ten-fold cross-validation on all the experiments and report mean and standard deviation on all classification metrics. To keep the experimental setup comparable across methods, we train all GAT experiments for 200 epochs, use the same learning rate (0.001) and hidden representations across experiments (50\% of its input modality dimension) when applicable. We performed our experiments using PyTorch and on a consumer-level GPU (Nvidia GTX 1080 Ti).

%
\section{Results}
\label{sec:res}
\begin{figure}[ht!]
  \centering
  \includegraphics[width=0.7\textwidth]{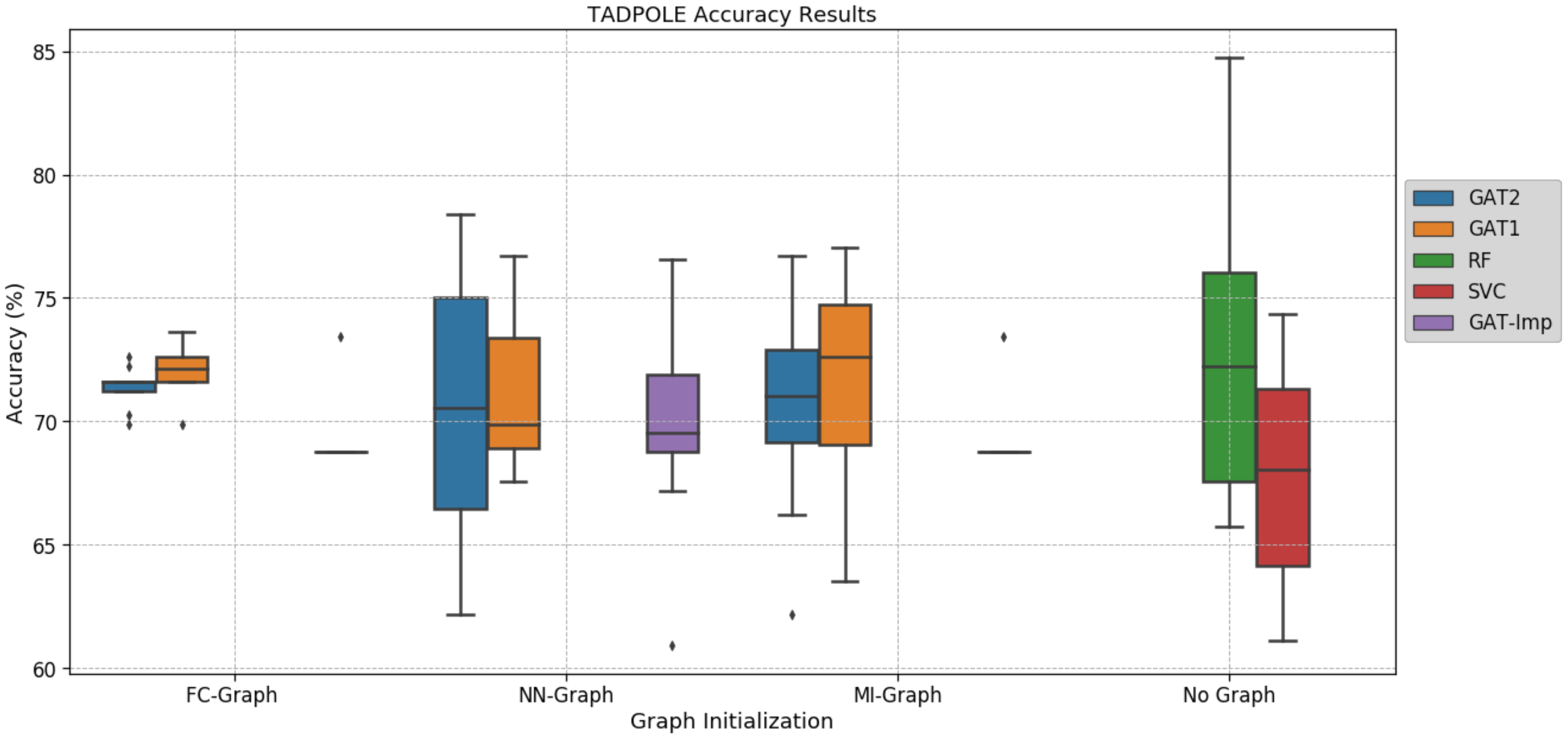}\\
  \caption{TADPOLE Accuracy Results.}
  \label{fig:TadpoleAccuracy}
\end{figure}

\begin{figure}[ht!]
  \centering
  \includegraphics[width=0.8\textwidth]{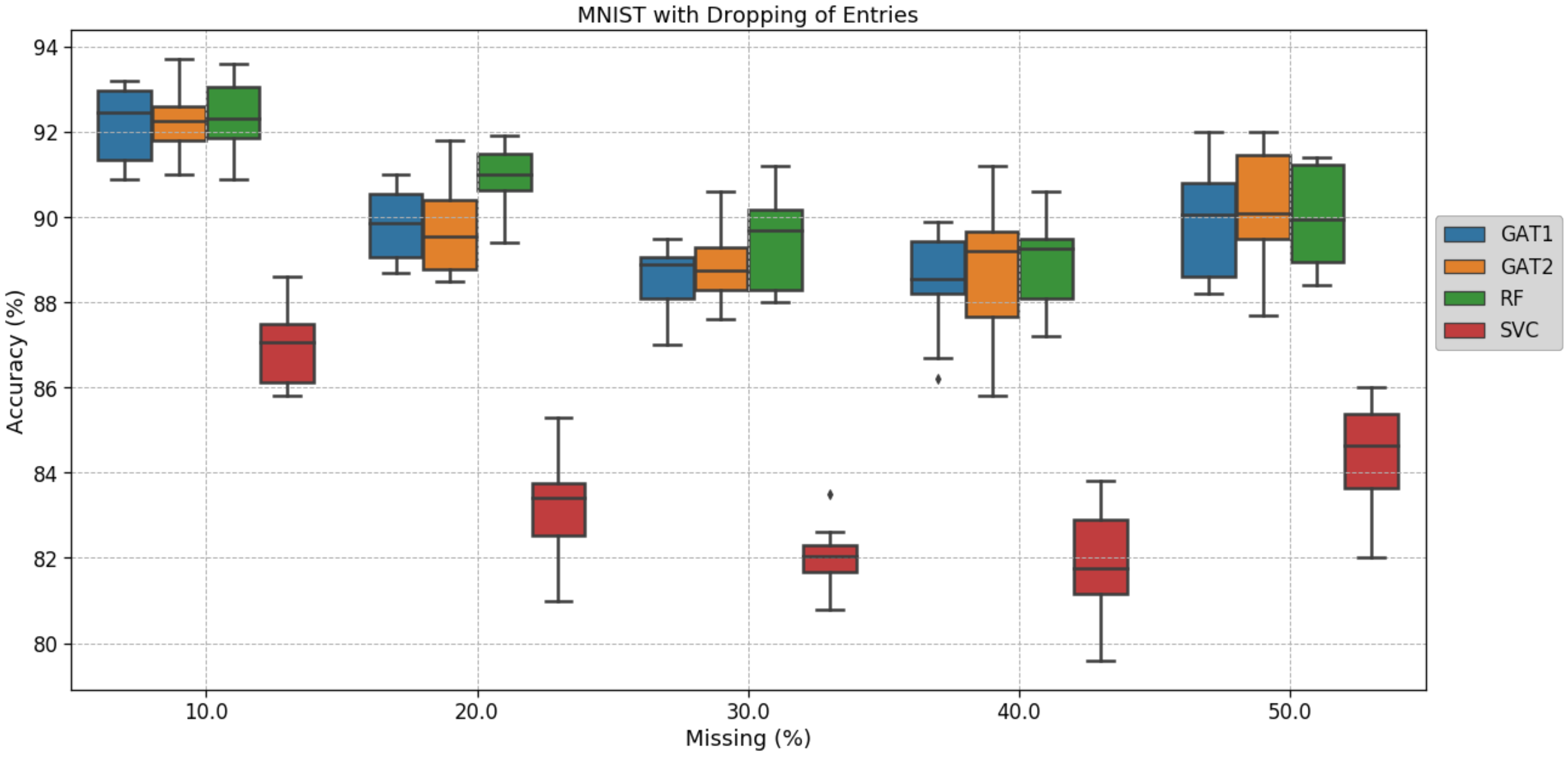}\\
  \caption{MNIST Accuracy Results.}
  \label{fig:MNISTResAcc}
\end{figure}

On the TADPOLE dataset, classification accuracies are illustrated in Figure \ref{fig:TadpoleAccuracy}, which ranged from 67.67\% to 72.64\%. 
GAT1 (72.07\%) performed second best to RF, while SVC has the lowest accuracy (67.68\%). With different graph initializations, GAT1 with MI-Graph and NN-Graph has the highest ROC, 0.569 and 0.563, respectively. On MNIST  (cf. Figure \ref{fig:MNISTResAcc}), accuracy scores ranged from 81.87\% to 92.37 \% with the NN-Graph initialization. With 10\% missing, RF performed the highest at 92.37 \% while SVC performed the lowest (86.98\%). With 50\% of the data missing (block-wise in three pseudo-modalities), GAT2 performed higher than RF with 90.26 \% accuracy. We did not include GAT-Imp in the figure due to its clearly inferior performance, with average scores ranging from 10\% to 24\%. 

\section{Discussion and Conclusion}
\label{sec:dis}
In this work, we propose a late fusion approach using multiple GAT to solve multi-modal disease prediction in incomplete datasets with block-wise or modality-wise missingness. Our initial results using this method on TADPOLE show that our GAT-Fusion approach works better than a standard GAT with mean imputation approach which uses a single graph. The single graph approach (GAT-Imp) works when a sparse Graph is used as initialization such as a NN-Graph and is very sensitive to the prior graph. In our propose GAT-Fusion approach, both GAT2 and GAT1 work better than the single graph approach as the network can leverage information from the other graphs. The reason behind this is that the network is able to pick-up useful information from other graphs via end-to-end training while a single static graph is heavily relying on a proper graph initialization. In contrast, our GAT-Fusion approach is more robust to graph initializations. It is important to note that our approach performs comparably, but cannot yet outperform a strong standard machine learning model like Random Forests, even with missing data that is mean-imputed. Nonetheless, we have shown that our proposed multiple graph setting works better than a single graph setting when performing node-level classification in this difficult setting of blockwise missingness. Furthermore, on MNIST experiments, GAT-Fusion approaches start to outperform standard machine learning models at 40\% and \%50 missingness, including standard GAT-Imp. The results on these experiments show the potential of our proposed GAT-Fusion approach, in case of multi-modal data with blockwise missing data, which is an important prerequisite for applying GAT approaches to realistic clinical datasets. In conclusion, we have shown that a multi-graph fusion approach to multi-modal missing could be a solution to this problem.
{
\bibliographystyle{plainnat}
\bibliography{Bibliography.bib}

\begin{thebibliography}{10}
\providecommand{\natexlab}[1]{#1}
\providecommand{\url}[1]{\texttt{#1}}
\expandafter\ifx\csname urlstyle\endcsname\relax
  \providecommand{\doi}[1]{doi: #1}\else
  \providecommand{\doi}{doi: \begingroup \urlstyle{rm}\Url}\fi

\bibitem[Bronstein et~al.(2017)Bronstein, Bruna, Lecun, Szlam, and
  Vandergheynst]{GCN:Bronstein17}
Michael~M. Bronstein, Joan Bruna, Yann Lecun, Arthur Szlam, and Pierre
  Vandergheynst.
\newblock {Geometric Deep Learning: Going beyond Euclidean data}.
\newblock \emph{IEEE Signal Processing Magazine}, 34\penalty0 (4):\penalty0
  18--42, 2017.

\bibitem[Bruna et~al.(2013)Bruna, Zaremba, Szlam, and LeCun]{Bruna2013}
Joan Bruna, Wojciech Zaremba, Arthur Szlam, and Yann LeCun.
\newblock {Spectral Networks and Locally Connected Networks on Graphs}.
\newblock pages 1--14, 2013.
\newblock URL \url{http://arxiv.org/abs/1312.6203}.

\bibitem[Defferrard et~al.(2016)Defferrard, Bresson, and
  Vandergheynst]{GCN:Defferrard16}
Micha{\"e}l Defferrard, Xavier Bresson, and Pierre Vandergheynst.
\newblock Convolutional neural networks on graphs with fast localized spectral
  filtering.
\newblock In \emph{Advances in Neural Information Processing Systems (NIPS)},
  pages 3844--3852, 2016.

\bibitem[Hamilton et~al.(2017)Hamilton, Ying, and Leskovec]{Hamilton2017}
William~L. Hamilton, Rex Ying, and Jure Leskovec.
\newblock {Inductive Representation Learning on Large Graphs}.
\newblock \penalty0 (Nips):\penalty0 1--19, 2017.
\newblock URL \url{http://arxiv.org/abs/1706.02216}.

\bibitem[Kipf and Welling(2016)]{Kipf2016}
Thomas~N. Kipf and Max Welling.
\newblock {Semi-Supervised Classification with Graph Convolutional Networks}.
\newblock pages 1--14, 2016.
\newblock ISSN 0004-6361.
\newblock \doi{10.1051/0004-6361/201527329}.
\newblock URL \url{http://arxiv.org/abs/1609.02907}.

\bibitem[Marinescu et~al.(2018)Marinescu, Oxtoby, Young, Bron, Toga, Weiner,
  Barkhof, Fox, Klein, Alexander, the EuroPOND~Consortium, and for~the
  Alzheimer's Disease Neuroimaging~Initiative]{Marinescu2018}
Razvan~V. Marinescu, Neil~P. Oxtoby, Alexandra~L. Young, Esther~E. Bron,
  Arthur~W. Toga, Michael~W. Weiner, Frederik Barkhof, Nick~C. Fox, Stefan
  Klein, Daniel~C. Alexander, the EuroPOND~Consortium, and for~the Alzheimer's
  Disease Neuroimaging~Initiative.
\newblock Tadpole challenge: Prediction of longitudinal evolution in
  alzheimer's disease.
\newblock 2018.

\bibitem[Parisot et~al.(2017{\natexlab{a}})Parisot, Ktena, Ferrante, Lee,
  Moreno, Glocker, and Rueckert]{GCN:Parisot17}
Sarah Parisot, Sofia~Ira Ktena, Enzo Ferrante, Matthew Lee, Ricardo~Guerrerro
  Moreno, Ben Glocker, and Daniel Rueckert.
\newblock Spectral graph convolutions for population-based disease prediction.
\newblock In \emph{International Conference on Medical Image Computing and
  Computer-Assisted Intervention}, pages 177--185, 2017{\natexlab{a}}.

\bibitem[Parisot et~al.(2017{\natexlab{b}})Parisot, Ktena, Ferrante, Lee,
  Moreno, Glocker, and Rueckert]{Parisot2017}
Sarah Parisot, Sofia~Ira Ktena, Enzo Ferrante, Matthew Lee, Ricardo~Guerrerro
  Moreno, Ben Glocker, and Daniel Rueckert.
\newblock {Spectral graph convolutions for population-based disease
  prediction}.
\newblock \emph{Lecture Notes in Computer Science (including subseries Lecture
  Notes in Artificial Intelligence and Lecture Notes in Bioinformatics)}, 10435
  LNCS\penalty0 (319456):\penalty0 177--185, 2017{\natexlab{b}}.
\newblock ISSN 16113349.
\newblock \doi{10.1007/978-3-319-66179-7_21}.

\bibitem[Thung et~al.(2016)Thung, Adeli, Yap, and Shen]{SWMC:Thung16}
Kim-Han Thung, Ehsan Adeli, Pew-Thian Yap, and Dinggang Shen.
\newblock Stability-weighted matrix completion of incomplete multi-modal data
  for disease diagnosis.
\newblock In \emph{International Conference on Medical Image Computing and
  Computer-Assisted Intervention}, pages 88--96, 2016.

\bibitem[Veli{\v{c}}kovi{\'{c}} et~al.(2017)Veli{\v{c}}kovi{\'{c}}, Cucurull,
  Casanova, Romero, Li{\`{o}}, and Bengio]{Velickovic2017}
Petar Veli{\v{c}}kovi{\'{c}}, Guillem Cucurull, Arantxa Casanova, Adriana
  Romero, Pietro Li{\`{o}}, and Yoshua Bengio.
\newblock {Graph Attention Networks}.
\newblock pages 1--12, 2017.
\newblock URL \url{http://arxiv.org/abs/1710.10903}.

\end{thebibliography}
}

\end{document}